\pdfoutput=1
\documentclass[11pt]{article}
\usepackage{CJKutf8}
\usepackage[preprint]{acl}
\usepackage{xcolor}
\usepackage{times}
\usepackage{latexsym}
\usepackage{booktabs}
\usepackage{amsfonts}
\usepackage{subcaption}
\usepackage{float}
\usepackage{pgfplots}
\usepackage[T1]{fontenc}
\usepackage[utf8]{inputenc}

\usepackage{microtype}

\usepackage{inconsolata}
\usepackage{color,soul}

\sethlcolor{flowblue!60}
\usepackage{graphicx}
\usepackage{amsmath}
\usepackage{tikz}

\usetikzlibrary{positioning, shapes, arrows.meta, decorations.pathreplacing, calc}
\definecolor{flowred}{RGB}{240, 130, 141}
\definecolor{flowyellow}{RGB}{244, 224, 138}
\definecolor{flowgreen}{RGB}{150, 225, 191}
\definecolor{flowblue}{RGB}{158, 150, 238}
\definecolor{myred}{RGB}{204, 0, 0}
\definecolor{mygreen}{RGB}{0, 153, 51}
\definecolor{bargreen}{RGB}{126, 217, 87}
\definecolor{barblue}{RGB}{56, 182, 255}
\definecolor{barred}{RGB}{255, 49, 49}
\definecolor{baryellow}{RGB}{255, 222, 89}
\title{Evaluating the Effectiveness of Universal Dependencies in Pretrained Language Models}
\title{Evaluating the Effectiveness of Linguistic Knowledge in Pretrained Language Models: A Case Study of Universal Dependencies}

\author{Wenxi Li \\
  School of the Chinese Nation Studies, Minzu University of China \\
  School of Liberal Arts, Minzu University of China \\
  \texttt{liwenxi@pku.edu.cn} \\}

\begin{document}
\maketitle
\begin{abstract}
Universal Dependencies (UD), while widely regarded as the most successful linguistic framework for cross-lingual syntactic representation, remains underexplored in terms of its effectiveness. This paper addresses this gap by integrating UD into pretrained language models and assesses if UD can improve their performance on a cross-lingual adversarial paraphrase identification task. Experimental results show that incorporation of UD yields significant improvements in accuracy and $F_1$ scores, with average gains of 3.85\% and 6.08\% respectively.
These enhancements reduce the performance gap between pretrained models and large language models in some language pairs, and even outperform the latter in some others.
Furthermore, the UD-based similarity score between a given language and English is positively correlated to the performance of models in that language. Both findings highlight the validity and potential of UD in out-of-domain tasks.
\end{abstract}
\begin{CJK*}{UTF8}{gbsn}
\section{Introduction}
\label{intro}
Universal Dependencies \citep[UD;][]{nivre-etal-2016-universal, nivre-etal-2020-universal} is a linguistic framework designed to provide consistent syntactic representations across languages.
By using dependencies to capture relations, UDs represent a fundamental worldview of how entities participate in events, i.e., who does what to whom and where/when. This makes UD feasible for representing cross-lingual data, as evidenced by its successful development of over 250 treebanks covering more than 150 human languages\footnote{\url{https://universaldependencies.org/}}.

\begin{figure}[htbp]
    \centering
    \scalebox{0.58}{

\begin{tikzpicture}[->, >=stealth, sibling distance=2.5cm, level distance=1.25cm]

% ----- EN1: Jim won against Tim -----
\begin{scope}
\tikzset{
  nodestyle/.style={draw, rounded rectangle, fill=flowblue!80, text centered},
   level 1/.style={sibling distance=2cm, level distance=0.9cm},
level 2/.style={sibling distance=3cm, level distance=1.15cm},
}
\node[nodestyle] (en1root) {\textit{won}}
  child { node[nodestyle] (en1jim) {\textit{Jim}} }
  child { node[nodestyle] (en1tim) {\textit{Tim}}
    child { node[nodestyle] (en1against) {\textit{against}} }
  };
\node[below=2.3cm of en1root] (en1text) {EN1: \textit{Jim won against Tim}};
\draw (en1root) -- (en1jim) node[near end, above left] {\footnotesize{\texttt{nsubj}}};
\draw (en1root) -- (en1tim) node[near end, above right] {\footnotesize{\texttt{obl}}};
\draw (en1tim) -- (en1against) node[near end, above left] {\footnotesize{\texttt{case}}};
\end{scope}

% ----- ZH1: 吉姆打败了蒂姆 -----
\begin{scope}[xshift=6cm]
\tikzset{
  nodestyle/.style={draw, rounded rectangle, fill=flowyellow!80, text centered},
  level 1/.style={sibling distance=2cm, level distance=1.5cm},
level 2/.style={sibling distance=3cm, level distance=1.2cm},
}
\node[nodestyle] (zh1root) {\textit{打败}\textsubscript{won against}}
  child { node[nodestyle] (zh1jim) {\textrm{吉姆}\textsubscript{Jim}} }
  child { node[nodestyle] (zh1tim) {\textrm{蒂姆}\textsubscript{Tim}} }
  child { node[nodestyle] (zh1le) {\textit{了}} };
\node[below=2.1cm of zh1root] (zh1text) {ZH1: \textit{吉姆打败了蒂姆}};
\draw (zh1root) -- (zh1jim) node[near end, above left] {\footnotesize{\texttt{nsubj}}};
\draw (zh1root) -- (zh1tim) node[near end, above left] {\footnotesize{\texttt{obl}}};
\draw (zh1root) -- (zh1le) node[near end, above right] {\footnotesize{\texttt{aux}}};
\end{scope}

% ----- EN2: Tim won against Jim -----
\begin{scope}[yshift=-4cm]
\tikzset{
  nodestyle/.style={draw, rounded rectangle, fill=flowblue!80, text centered},
   level 1/.style={sibling distance=2cm, level distance=0.9cm},
level 2/.style={sibling distance=3cm, level distance=1.15cm},
}
\node[nodestyle] (en2root) {\textit{won}}
  child { node[nodestyle] (en2tim) {\textit{Tim}} }
  child { node[nodestyle] (en2jim) {\textit{Jim}}
    child { node[nodestyle] (en2against) {\textit{against}} }
  };
\node[below=2.3cm of en2root] (en2text) {EN2: \textit{Tim won against Jim}};
\draw (en2root) -- (en2tim) node[near end, above left] {\footnotesize{\texttt{nsubj}}};
\draw (en2root) -- (en2jim) node[near end, above right] {\footnotesize{\texttt{obl}}};
\draw (en2jim) -- (en2against) node[near end, above left] {\footnotesize{\texttt{case}}};
\end{scope}

% ----- ZH2: 蒂姆打败了吉姆 -----
\begin{scope}[xshift=6cm, yshift=-4cm]
\tikzset{
  nodestyle/.style={draw, rounded rectangle, fill=flowyellow!80, text centered},
   level 1/.style={sibling distance=2cm, level distance=1.5cm},
level 2/.style={sibling distance=3cm, level distance=1.2cm},
}
\node[nodestyle] (zh2root) {\textit{打败}\textsubscript{won against}}
  child { node[nodestyle] (zh2tim) {\textrm{蒂姆}\textsubscript{Tim}} }
  child { node[nodestyle] (zh2jim) {\textrm{吉姆}\textsubscript{Jim}} }
  child { node[nodestyle] (zh2le) {\textit{了}} };
\node[below=2.1cm of zh2root] (zh2text) {ZH2: \textit{蒂姆打败了吉姆}};
\draw (zh2root) -- (zh2tim) node[near end, above left] {\footnotesize{\texttt{nsubj}}};
\draw (zh2root) -- (zh2jim) node[near end, above left] {\footnotesize{\texttt{obl}}};
\draw (zh2root) -- (zh2le) node[near end, above right] {\footnotesize{\texttt{aux}}};
\end{scope}

% ----- Arrows between trees -----
\coordinate (A) at ($(en1root)+(1.8,-0.7)$);
\coordinate (B) at ($(en2root)+(1.8,-1.1)$);
\coordinate (C) at ($(zh1root)+(-2.7,-0.7)$);
\coordinate (D) at ($(zh2root)+(-2.7,-1.1)$);

% Red arrows (EN1 <-> EN2 and ZH1 <-> ZH2)
\draw[color=myred, -{Latex[scale=1]}, ultra thick] (A) -- (D);
\draw[color=myred, -{Latex[scale=1]}, ultra thick] (D) -- (A);
\draw[color=myred, -{Latex[scale=1]}, ultra thick] (C) -- (B);
\draw[color=myred, -{Latex[scale=1]}, ultra thick] (B) -- (C);

% Green arrows (EN1 <-> ZH1 and EN2 <-> ZH2)
\draw[color=mygreen, -{Latex[scale=1]}, ultra thick] (A) -- (C);
\draw[color=mygreen, -{Latex[scale=1]}, ultra thick] (C) -- (A);
\draw[color=mygreen, -{Latex[scale=1]}, ultra thick] (B) -- (D);
\draw[color=mygreen, -{Latex[scale=1]}, ultra thick] (D) -- (B);

\end{tikzpicture}
}

\caption{Cross-lingual sentence pairs which are semantically aligned at the lexical level. Green and red arrows indicate that they are paraphrased or not respectively.}
\label{fig:intro}
\end{figure}
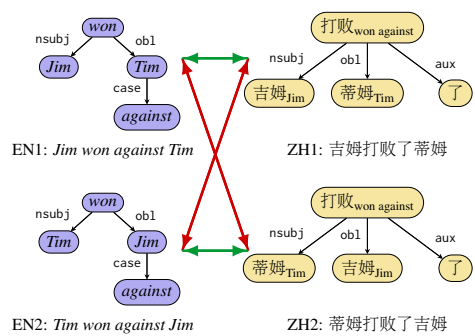

While UD has become a leading framework for cross-lingual syntactic representations, most research has focused on its annotation, parsing, and evaluation \citep{mcdonald-etal-2013-universal,qi-etal-2018-universal,nivre-fang-2017-universal}, with relatively little attention given to its grounding in other out-of-domain tasks.
To address the gap, this paper introduces a cross-lingual adversarial paraphrase identification (PI) task.
Adversarial examples of the PI task are sentences which share lexical overlap but differ significantly in semantics \citep{zhang-etal-2019-paws,yang-etal-2019-paws}, posing a major challenge for pre-trained language models (LMs).
We argue that the same situation persists in a cross-lingual context.
As shown in Figure \ref{fig:intro}, some cross-lingual sentence pairs exhibit high-degree lexical alignment but overall do not qualify as paraphrases.
In our view, these cross-lingually adversarial examples underscore the necessity for modeling their syntactic similarities across languages, which thus could be an ideal testing ground to evaluate the effectiveness of UD. 

We therefore explore to integrate UD into pretrained language models (PLMs; \textsection \ref{method}) and evaluate the performance of UD-enhanced models using the PAWS-X dataset \citep[][\textsection \ref{experiment}]{yang-etal-2019-paws}. 
Our experiments show that leveraging syntactic similarities across languages captured by UD, improves PLMs performance on the cross-lingual adversarial PI task, making them competitive with large language models (LLMs). 
Furthermore, the calculated similarity scores offer predictive insights into model performance across language pairs. Together, these findings underscore the effectiveness of UD. 
\section{UD-enhanced Models}
\label{method}
The following introduces how we explicitly integrate UD's cross-lingual representations to the self-attention mechanisms of pretrained models.
\subsection{Transforming Dependencies into Hypergraphs}
\label{construction}
We begin by converting the dependency structure of UD into a hypergraph, a generalization of a graph where edges (hyperedges) can connect multiple nodes (hypernodes) simultaneously, to facilitate the integration of UD into the system.
Specifically, each word in the UD representation, along with its relation to the head, forms a hypernode. The corresponding hyperedge, directing towards it, connects this hypernode to its set of dependent hypernodes.

More formally, a hypergraph-based dependency can be represented as a pair $\langle V, E \rangle$, where $V$ is the set of hypernodes, and $E$ is the set of hyperedges. For a given sentence $w_{1:l
} = w_1 \dots w_l$, each hypernode $v_l \in V$ takes the form $w_{l\_\text{label}}$, indicating that $w_l$ holds a syntactic relation labeled as \textit{label} with its head.
Each hyperedge $e_l \in E$ is defined as a tuple $\langle \langle \text{dependents}(v_l) \rangle, v_l \rangle$, where the hyperedge directs to $v_l$ and $\langle \text{dependents}(v_l) \rangle$ includes all of its dependent hypernodes. By allowing the set of dependents to be empty, we assume that every node $v_l \in V$ can function as a head, facilitating later comparisons.

Take the EN1 and ZH1 sentences in \textsection \ref{intro} as an example, their corresponding hyperedges can be represented as below: 
\begin{table}[htbp]
    \centering
    \scalebox{0.77}{
    \begin{tabular}{ccc}\toprule
        Sen &ID & Hyperedge \\\midrule
        EN1&$e_0$  & $\langle \langle \rangle, \textrm{Jim}_{\textrm{nsubj}}\rangle$ \\
        EN1&$e_1$ & $\langle \langle \textrm{Jim}_{\textrm{nsubj}}, \textrm{Tim}_{\textrm{obl}}\rangle, \textrm{won}_{\textrm{root}}\rangle$ \\
    EN1&$e_2$ & $\langle \langle \rangle, \textrm{against}_{\textrm{case}}\rangle$ \\
    EN1&$e_3$ & $\langle \langle \textrm{against}_{\textrm{case}} \rangle, \textrm{Tim}_{\textrm{obl}}\rangle$ \\
     ZH1&$e_0$& $\langle \langle \rangle, \textrm{吉姆}_{\textrm{nsubj}}\rangle$ \\
     ZH1&$e_1$ & $\langle \langle \textrm{\textit{吉姆}}_{\textrm{nsubj}}, \textrm{蒂姆}_{\textrm{obj}}, \textrm{了}_{\textrm{aux}}\rangle, \textrm{打败}_{\textrm{root}}\rangle$  \\
     ZH1&$e_2$ & $\langle \langle \rangle, \textrm{了}_{\textrm{aux}} \rangle$  \\
     ZH1&$e_3$ & $\langle \langle \rangle, \textrm{蒂姆}_{\textrm{obj}} \rangle$ \\
      \bottomrule
    \end{tabular}}
    \caption{Dependency structures represented with hyperedges in English and Chinese.}
\label{fig:hypergraph_example}
\end{table}

% This span information is essential for distinguishing among nodes and indicating the scope of heads by clearly showing the words they dominate. For instance, in the first sentence, the word ``a'' occurs twice, once with the span (2, 3) and once with the span (5, 6). The fourth word ``boy,'' which dominates the node \(a_{2,3}\), can be referred to as \(boy_{2,4}\) to indicate its hierarchical position and scope.

We believe that, unlike dependency trees which use directed edges to represent relationships between heads and dependents, hypergraphs capture higher-order syntactic dependencies by grouping dependents with a common head into hyperedges. This structure preserves the integrity of substructures, avoiding the branch-wise fragmentation typical of tree-based representations.

\subsection{Constructing Hypergraph-based Similarity Matrix}
We then construct a similarity matrix by comparing the hypergraphs (see more related work in Appendix \ref{related}). Specifically, for two sentences of lengths $n$ and $m$, with their respective hypergraphs $G_A$ and $G_B$, we index the hyperedges of $G_A$ as $e_i$ for $i \in \{0, \dots, n-1\}$ and those of $G_B$ as $e_j$ for $j \in \{0, \dots, m-1\}$. The similarity matrix $M \in \mathbb{R}^{n \times m}$ is then defined as:
\[
\scalebox{0.89}{$
\begin{aligned}
M_{ij} = \operatorname{Sim}(e_i, e_j)
\end{aligned}
$}
\]
where the $Sim$ function compares the two hyperedges based on the lexical alignment of their hypernodes and the similarity of the labels, and then adjusts the weights accordingly. Further details are provided below.

% this approach aims to effectively incorporate hypergraph-based cross-lingual syntactic information into the system.

% comparing hyperedges within the constructed hypergraphs and implementing weight updates, the similarity scores between each pair of tokens in two sentences are computed, which is a constituent of the whole matrix.

\paragraph{Comparison of Hypernodes}
The comparison function $Sim_N$ between two hypernodes $v_i$, $v_j$, which are represented by $w_{i\_\text{label}_i}$ and $w_{j\_\text{label}_j}$, consists of two components: word alignment and label comparison.
For word alignment, we utilize the \textit{SimAligner} \citep{simalign} as it is a lightweight yet effective tool. Specifically, it leverages the multilingual BERT model (mBERT)\footnote{\url{https://github.com/google-research/bert/blob/master/multilingual.md}}, which supports 104 languages, to generate multilingual embeddings for target tokens, and further uses IterMax, a heuristic algorithm that adopts a greedy approach, allowing a single token to be aligned with multiple others.
For label comparison, we assess the equivalence of two labels. Consequently, $Sim_N$ can be described as follows:
\[
\scalebox{0.89}{$
\begin{aligned}
Sim_{N}(v_i, v_j) &=\underbrace{s\left(w_i, w_j\right)}_{\text{word alignment}} \times \underbrace{q\left(\text{label}_i, \text{label}_i\right)}_{\text {label comparison}}  \\
s(w_i, w_j) &= 
\begin{cases} 
1 & \text{if } w_i \text{ aligns with } w_j \\ 
0 & \text{otherwise} 
\end{cases} \label{eq:s} \\
q(\text{label}_i, \text{label}_j) &= 
\begin{cases} 
\theta & \text{if } \text{label}_i = \text{label}_j \\ 
1 & \text{otherwise} 
\end{cases} 
\end{aligned}$}
\]

\paragraph{Comparison of Hyperedges}
% The novel hypergraph-based kernel function introduced in this study commences by iteratively examining and comparing pairs of hyperedges, denoted as $e_A^i$ and $e_B^j$, extracted from the two hypergraphs $G_A$ and $G_B$, as constructed according to the methodology outlined in \textsection \ref{construction}. Here, $e_A^i$ signifies the $i_{th}$ hyperedge in hypergraph $G_A$, which is in the form $\langle \langle d_A^1\_{t_A^1} \dots d_A^k\_{t_A^k} \rangle, h_A^i\_{t_A^i}\rangle$.

The comparison of hyperedges $e_i$, $e_j$, formed as $\langle \langle \text{dependents}(v_i) \rangle, v_i \rangle$ and $\langle \langle \text{dependents}(v_j) \rangle, v_j \rangle$, involves two steps. 
In the first step, we compare the head nodes $v_i$, $v_j$ using he function $Sim_{N}$.
If the similarity score between the heads is non-zero, the $Sim_{N}$ function is then iteratively applied to compare their dependent nodes.
We index the dependents of $v_i$ as $d_k$ for $k \in \{0, \dots, k-1\}$ and those of $v_j$ as $d_l$ for $l \in \{0, \dots, l-1\}$ , where $k$ and $l$ are lengths of the dependent sets of the two hyperedges respectively. The outcomes of these comparisons of dependent nodes are cumulatively aggregated and subsequently multiplied by the similarity score of the head nodes.
So the function for comparing the similarity between the two hyperedges, $Sim_{E}$, is defined as follows:
\[
\scalebox{0.89}{$
\begin{aligned}
 Sim_{E}(e_i, e_j) = & \underbrace{Sim_{N}({v_i}, {v_j})}_{\text {head node}} \times \underbrace{\sum_{k=0}^{k-1} \sum_{l=0}^{l-1} Sim_{N}({d_k}, {d_l})}_{\text {dependent node}}   
\end{aligned}$}
\]

\paragraph{Update of Weights}
Intuitively, the weights of different hyperedges, which indicate their contributions to determining the degree of similarity between two hypergraphs, exhibit disparities. This recognition arises from the fact that the hyperedge headed by the root node, which represents the main verb in a sentence, is of paramount importance. It embodies the fundamental structure of the sentence, namely who does what to whom, and when or where the event occurs. To capture this aspect, we calculate the height of each node to derive the weights of their corresponding hyperedges, ultimately leading to the similarity function $Sim(e_i, e_j)$ between two hyperedges.
Formally, the function is defined as follows:
\[
\scalebox{0.89}{$
\begin{aligned}
Sim(e_i, e_j)=Sim_{E}(e_i, e_j) \times h(v_i) \times h(v_j)
\end{aligned}$}
\]

Here, $h(v_i)$, $h(v_j)$ represent the heights of the two nodes in the dependency tree, which serve as the weights of their corresponding hyperedges. The calculation of heights is performed using the $h(v)$ function, where the heights of leaf nodes are set to 1 while those of other nodes are determined by the maximum height between their left and right child nodes $v_l$ and $v_r$, with an augmentation of $\beta$.
\[
\scalebox{0.89}{$
\begin{aligned}
h(v)=
\begin{cases} 
1 & \text{if } v \text{ is a leaf node} \\ 
\max(h(v_l), h(v_r)) + \beta & \text{otherwise} 
\end{cases}
\end{aligned}
$}
\]

\subsection{Injecting the Similarity Matrix}
We further integrate the derived similarity matrix into the attention mechanism. This enables the attention module to focus more effectively on syntactically relevant relationships between the sentences, refining the attention distributions.

More specifically, after the matrix is appropriately padded or truncated, the UD-aware attention score for each head is element-wise multiplied by it, as shown in the following equation. Here, $M$ represents the matrix, while $\mathcal{Q}$, $\mathcal{K}$, and $\mathcal{V}$ denote the query, key, and value matrices, respectively. The term $d$ stands for the dimension of $\mathcal{K}$.
\[
\scalebox{0.89}{$
\begin{aligned}
\operatorname{UDAtt}\left(\mathcal{Q},\mathcal{K}, \mathcal{V}, M\right)=\operatorname{softmax}\left(\frac{\mathcal{Q K}^\top \odot {M}}{\sqrt{d}}\right) \times \mathcal{V}
\end{aligned}
$}
\]

\section{Experiment}
\label{experiment}
\subsection{Data Preparation}
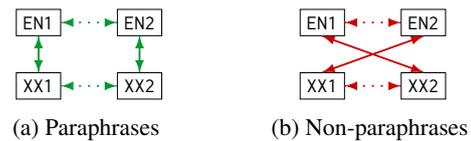
\begin{figure}[ht]
    \centering
    \begin{subfigure}[b]{0.45\columnwidth}
        \centering
        \scalebox{0.7}{
        \begin{tikzpicture}[node distance=1.2cm]
            % Define nodes
            \node (en1) [draw] {\texttt{EN1}};
            \node (en2) [draw, right of=en1, xshift=0.7cm] {\texttt{EN2}};
            \node (nonen1) [draw, below of=en1] {\texttt{XX1}};
            \node (nonen2) [draw, below of=en2] {\texttt{XX2}};

            % Draw arrows
            \draw[color=mygreen, loosely dotted, -{Latex[scale=1]}, thick] (nonen1) -- (nonen2) node[midway, below] {};
            \draw[color=mygreen,loosely dotted, -{Latex[scale=1]}, thick] (nonen2) -- (nonen1);
            \draw[color=mygreen,loosely dotted, -{Latex[scale=1]}, thick] (en1) -- (en2) node[midway, above] {};
            \draw[color=mygreen,loosely dotted, -{Latex[scale=1]}, thick] (en2) -- (en1);
            \draw[color=mygreen,-{Latex[scale=1]}, thick] (nonen1) -- (en1) node[midway, left] {};
            \draw[color=mygreen,-{Latex[scale=1]}, thick] (en1) -- (nonen1);
            \draw[color=mygreen,-{Latex[scale=1]}, thick] (nonen2) -- (en2) node[midway, right] {};
            \draw[color=mygreen,-{Latex[scale=1]}, thick] (en2) -- (nonen2);
        \end{tikzpicture}}
        \caption{Paraphrases}
    \end{subfigure}
    \hspace{0.01\columnwidth} % Space between the figures
    \begin{subfigure}[b]{0.45\columnwidth}
        \centering
        \scalebox{0.7}{
        \begin{tikzpicture}[node distance=1.2cm]
            % Define nodes
            \node (nonen1) [draw] {\texttt{EN1}};
            \node (nonen2) [draw, right of=nonen1, xshift=0.7cm] {\texttt{EN2}};
            \node (en1) [draw, below of=nonen1] {\texttt{XX1}};
            \node (en2) [draw, below of=nonen2] {\texttt{XX2}};

            % Draw arrows
            \draw[color=myred, loosely dotted, -{Latex[scale=1]}, thick] (nonen1) -- (nonen2) node[midway, above] {};
            \draw[color=myred,loosely dotted, -{Latex[scale=1]}, thick] (nonen2) -- (nonen1);
            \draw[color=myred,loosely dotted, -{Latex[scale=1]}, thick] (en1) -- (en2) node[midway, below] {};
            \draw[color=myred,loosely dotted, -{Latex[scale=1]}, thick] (en2) -- (en1);
            \draw[color=myred,-{Latex[scale=1]}, thick] (nonen1.south) -- (en2.north) node[pos=0.5, left, xshift=-0.5cm] {};
            \draw[color=myred,-{Latex[scale=1]}, thick] (en2.north) -- (nonen1.south);
            \draw[color=myred,-{Latex[scale=1]}, thick] (nonen2.south) -- (en1.north) node[pos=0.5, right, xshift=0.5cm] {};
            \draw[color=myred,-{Latex[scale=1]}, thick] (en1.north) -- (nonen2.south);
        \end{tikzpicture}}
        \caption{Non-paraphrases}
    \end{subfigure}
    \caption{Sentence pair reorganization. Dotted arrows show original pairings while the solid indicate new ones.\label{generation}}
\end{figure}

%%%%%%%%data set
We use the PAWS-X dataset \cite{yang-etal-2019-paws} as our experimental data source. PAWS-X, an extension of the PAWS dataset, contains adversarial sentence pairs in seven languages: English, French, Spanish, German, Chinese, Japanese, and Korean. Since all non-English instances are translations of their English counterparts, the dataset is well-suited for the cross-lingual adversarial PI task.
We focus on the human-translated sentences from the development set of PAWS-X. This selection follows the suggestion in \citet{yang-etal-2019-paws} as the training set is machine-translated, and the test set contains some duplicated sentences. After filtering out samples with missing translations, we retain 1848 sentence pairs for each language. As illustrated in Figure \ref{generation}, these are then reorganized cross-lingually into new pairs. Finally, all sentence pairs in the newly-created dataset are parsed into UD-style dependencies using the Stanza parser \citep{qi2020stanza}.

\subsection{Models}
We use BERT-base-multi and BERT-large-multi \cite{devlin-etal-2019-bert}, XLM-RoBERTa-base and XLM-RoBERTa-large \citep{conneau-etal-2020-unsupervised} as our baselines. They are enhanced using the method described in \textsection\ref{method}, resulting in their UD-aware counterparts. Llama 3 \citep{grattafiori2024llama3herdmodels}, which is the best-performing LLM in a classification task \citep{ruan-etal-2024-large}, is employed as a reference.
Detailed implementation of them is provided in Appendix \ref{sec:appendix-detail}. 
\begin{table*}[h]
    \centering
    \scalebox{0.88}{
     % \begin{tabular}{l p{0.7cm} p{0.7cm} p{0.7cm} p{0.7cm} p{0.7cm} p{0.7cm}}
     \begin{tabular}{lcccccc}
     \toprule
         Model  & EN-FR& EN-ES& EN-DE& EN-ZH& EN-JA& EN-KO\\\midrule\midrule
          Bert-base& 68.11/64.02  &66.76/64.76 & 71.08/68.06 & 60.54/49.66 & 62.16/59.77 & 63.78/55.92\\
          UD-Bert-base&73.51/70.83 &70.00/68.56 & 74.59/72.67&66.49/58.94 & 66.76/60.70 &67.03/64.74\\\midrule
          Bert-large& 71.35/61.59& 70.81/64.71& 72.16/65.32&64.32/46.34 &65.41/55.56 &68.11/59.869\\
          UD-Bert-large&78.38/75.16 &72.97/66.44 & 79.19/76.01&67.30/61.09 &68.92/60.48 &70.81/65.16\\\midrule
          Roberta-base& 80.81/79.42 & 73.78/73.13& 78.65/75.84&64.59/52.01 & 67.57/59.46&66.49/59.21 \\
          UD-Roberta-base& 87.84/86.57&77.57/77.75 & 84.59/82.57& 68.11/61.69&69.73/60.56 &69.73/64.33\\\midrule
          Roberta-large&91.08/90.21 &88.92/87.91 &89.18/87.65 & 66.22/59.81& 64.32/52.52&68.92/62.78\\
          UD-Roberta-large&95.41/94.64 &93.51/92.64 &90.27/89.02&66.76/61.20 & 64.85/61.64& 69.73/68.18\\\midrule
          Llama3-8B-Instruct & 90.86/89.86 & 88.65/89.35 & 88.11/89.34 & 77.30/78.94 &74.57/75.60 & 78.26/78.19 \\
          \bottomrule
    \end{tabular}}
    \caption{Performance of baseline PLMs, their UD-enhanced variants, and an LLM on the test set.\label{result}}
\end{table*}
% {'accuracy': 0.6486486486486487, 'f1': 0.6264367816091955}
% % roberta-base-fr-without-ud {'accuracy': 0.6675675675675675, 'f1': 0.6119873817034701}
% roberta-large for zh without ud {'accuracy': 0.6621621621621622, 'f1': 0.5980707395498391} 63.78/52.14x
% roberta-large for ko without ud 0.6972972972972973, 'f1': 0.6818181818181819}
% roberta-base for ja without ud {'accuracy': 0.6837837837837838, 'f1': 0.597938144329897}
\subsection{Result}
Experimental results, presented in Table~\ref{result}, demonstrate that incorporating UD information consistently improves both accuracy and $F_1$ scores of PLMs, with average gains of 3.85\% and 6.08\%, respectively. These gains are visualized in Appendix~\ref{sec:appendix-figure}.
We argue that the results testify UD’s effectiveness in capturing structural information \citep{liu-etal-2020-sentence,xu-etal-2022-semantic}.

In addition, comparing to the LLM which showcases its robust cross-lingual generalization, it can also be observed that PLMs display greater performance variance across languages pairs. They excel with Indo-European language pairs but underperform with others. We claim that this provide a promising direction for optimization --- enhancing generalization powers across languages to close the gap with, or even surpass, LLMs.

\section{Converting Matrices to Scalar Values}
% \subsubsection{Correlations}
% \subsubsection{From Matrix to Scalar}
We then compute similarity scores between cross-lingual sentence pairs by converting the constructed matrix $M \in \mathbb{R}^{n \times m}$ into a scalar value. The average similarity scores of sentences in different language pairs (see Table \ref{tab:score}), in our view, could be interpreted as an approximation of the syntactic distance between English and other languages\footnote{To minimize variations of different dependency structures, we compare English-English pairs and then normalize resulting scores accordingly.}. 
\[
\scalebox{0.89}{$
\begin{aligned}
Score(M)=\frac{\sum_{i=0}^{n-1} \sum_{j=0}^{m-1} Sim({e_i}, {e_j})}{m + n}  
\end{aligned}
$}
\]

\begin{table}[htbp]
    \centering
    \scalebox{0.84}{
    \begin{tabular}{ccccccc}\toprule
 & FR  & ES &DE &ZH &JA & KO  \\\midrule
Mean  & 0.949 & 0.796 &0.747 & 0.533&0.362& 0.526\\ 
SD   & 0.111   & 0.156 & 0.182& 0.170& 0.141&0.180 \\\bottomrule
    \end{tabular}}
    \caption{Descriptive statistics of similarity scores between non-English languages and English.}
    \label{tab:score}
\end{table}

As shown in Table~\ref{tab:correlation}, the similarity scores exhibit a positive correlation with the accuracy of all models, suggesting that cross-lingual divergence --- captured by UD-based measures --- can serve as a reliable predictor of model performance. 
It is also noteworthy that RoBERTa-large and LLaMA 3, representing the strongest PLM and LLM in our experiments, display the highest correlations. This observation implies that as models become more capable, their performance may converge to the bound set by linguistic divergence.

\begin{table}[htbp]
    \centering
    \scalebox{0.88}{
    \begin{tabular}{ccc}\toprule
 & Pearson & $p$-value  \\\midrule
Bert-base  & 0.766 & 0.076 \\ 
Bert-large   & 0.840   & 0.036\\
Roberta-base   & 0.871   & 0.024\\
Roberta-large   & 0.943   & 0.005\\
Llama3-8B-Instruct   & 0.973   & 0.001\\
\bottomrule
    \end{tabular}}
    \caption{The statistical values of  correlations between similarity scores and model performance.}
    \label{tab:correlation}
\end{table}

% Mean  & 0.949 & 0.111 & & \\ 
% SD   & 0.796   & 0.156 & &\\
% EN-DE     & 0.747  & 0.182 & &\\ 
% EN-ZH     & 0.533  & 0.170 & & \\
% EN-JA   & 0.362 &  0.141& &\\
% EN-KO     & 0.526  & 0.180 & & \\
\section{Conclusion}
% We argue that this paper firstly contributes to UD community --- by introducing a hypergraph-based approach for modelling syntactic similarities, we are enabled to validate effectiveness or applicability of UD representations in other NLP tasks. For adversarial PI task, whether they are cross-lingual or not, this paper offers a practical technique to enhance LMs by better capturing syntactic structure.

By injecting UD information into language models, this paper makes contributions in two key aspects. First, it advances the UD community by demonstrating UD’s effectiveness and broader applicability in downstream NLP tasks. 
Second, the paper benefits language models by showing that the incorporation of linguistically informed knowledge can yield practical performance gains and offer insights into optimization, which highlights the continued relevance of linguistics in the era of LLMs.

% offers a practical technique for enhancing language models in the cross-lingual adversarial PI tasks.
\end{CJK*}
\section*{Limitations}
This paper employs one LLM as a reference for comparison. It does not explore the effectiveness of UD in LLMs or how UD information could be leveraged for fine-tuning them. Investigating this remains an important direction for our future work.

\bibliography{custom}

\appendix
\section{Related Work: Dependency Kernels for Sentence Similarity Scores}
\label{appendix-related}
\citet{ozates-etal-2016-sentence} propose an approach that leverages dependency grammar representations to calculate sentence similarity for extractive multi-document summarization. By representing dependencies as bigrams, in the form {\textit{head}, \textsc{label}, \textit{dependent}}, they introduce a set of innovative dependency grammar-based kernels designed to capture similarities between sentences.

The basic version, called the \textit{Simple Approximate Bigram Kernel} (SABK), computes syntactic similarity between two sentences, A and B, by iterating over and comparing each bigram, $A_b{ }^i$ and $B_b{ }^j$ --- where $A_b{ }^i$ and $B_b{ }^j$ are bigrams with the $i_{th}/j_{th}$ word in sentence A or B as the tail node, respectively. For sentences of lengths $m$ and $n$, the similarity is established as follows:
\[
\scalebox{0.85}{$
\begin{aligned}
 S A B K(A, B)=\frac{\sum_{i=1}^m \sum_{j=1}^n \operatorname{sim}\left(A_b{ }^i, B_b{ }^j\right)}{m+n}  
\end{aligned}
$}
\]
More specifically, the comparison between the bigrams $A_b{ }^i (h_A^i, t_A^i, d_A^i)$ and $B_b{ }^j (h_B^j, t_B^j, d_B^j)$ involves analyzing their heads, dependents, and type nodes using the $s$ and $q$ functions, as shown in the following equations.
\[
\scalebox{0.85}{$
\begin{aligned}
    \operatorname{sim}&\left(A_b{ }^i, B_b{ }^j\right) \\
    &= \left[s\left(d_A^i, d_B^j\right)+s\left(h_A^i, h_B^j\right)\right] \times q\left(t_A{ }^i, t_B{ }^j\right) \\
     &s(a, b)=\{1 \text { or } 0 \mid \text { if } a=b \text { or } \text { otherwise }\} \\
 &q(a, b)=\{\theta \text { or } 1 \mid \text { if } a=b \text { or } \text { otherwise }\}
\end{aligned}
$}
\]

% \begin{equation}
% \begin{aligned}
% & s(a, b)= \begin{cases}1, & \text { if } a=b \\
% 0, & \text { otherwise }\end{cases} \\
% & q(a, b)= \begin{cases}\theta, & \text { if } a=b \\
% 1, & \text { otherwise }\end{cases}
% \end{aligned}

% \end{equation}

The authors also argue that not all bigrams in a dependency graph hold equal significance. To account for this, they integrate term frequency-inverse document frequency (tf-idf) values, as introduced by \citet{ramos2003using}, to measure the informativeness of individual bigrams. This leads to the development of the \textit{TF-IDF Based Approximate Bigram Kernel} (TABK).
\[
\scalebox{0.85}{$
\begin{aligned}
    T A B K(A, B)=\frac{\sum_{i=1}^m \sum_{j=1}^n\operatorname{sim_t}\left(A_b{ }^i, B_b{ }^j\right)}{N(A) \times N(B)}
\end{aligned}
$}
\]
In this refined kernel function, the tf-idf weights of the head and dependent tokens are multiplied with the original results, placing greater emphasis on key dependencies within the sentence.

% \[
% \scalebox{0.85}{$
% \begin{aligned}
%     \operatorname{sim}_t&\left(A_b^i, B_b^j\right)= \\
%              &\left[ \left( t f_{d_A^i}  \times t f_{d_B^j} \right) \times s\left(d_A^i, d_B^j\right) + \left(t f_{h_A^i} \times t f_{h_B^j} \right)\\
%              & \times s\left(h_A{ }^i, h_B{ }^j \right) \right] \times q \left( t_A{ }^i, t_B{ }^j \right) 
% \end{aligned}
% $}
% \]

\[
\scalebox{0.85}{$
\begin{aligned}
    \operatorname{sim}_t\left(A_b^i, B_b^j\right) = 
    &\left[ \left( t f_{d_A^i} \times t f_{d_B^j} \right) \times s\left(d_A^i, d_B^j\right) \right. \\
    &\left. + \left( t f_{h_A^i} \times t f_{h_B^j} \right) \times s\left(h_A^i, h_B^j \right) \right] \\
    &\times q\left(t_A^i, t_B^j \right)
\end{aligned}
$}
\]

The resulting value is then normalized using the normalizer function $N(A)$.
\[
\scalebox{0.85}{$
\begin{aligned}
N(A)=\sqrt{\sum_{i=1}^n\left(t f_{d_A i} i d f_{d_A i}\right)^2+\left(t f_{h_A i} i d f_{h_A}\right)^2}
\end{aligned}
$}
\]

In addition to comparing individual bigram units, the authors introduce the \textit{Matching Subtrees Kernel} (MSK), which examines consecutive dependency subtrees. As shown by the equation below, it recursively analyzes the $K/L$ child nodes ($c_{d_A{ }^i}(k), c_{d_B{ }^j}(l)$) of dependent nodes $d_A{ }^i$ and $d_B{ }^j$ within a matching bigram pair $A_b{ }^i$ and $B_b{ }^j$, using a Children Kernel ($K_c$). 
\[
\scalebox{0.85}{$
\begin{aligned}
     M S&K(A, B)= T A B K(A, B) + \\
     &\frac{\sum_{k=1}^K \sum_{l=1}^L s\left(d_A{ }^i, d_B{ }^j\right) \times K_c\left(c_{d_A{ }^i}(k), c_{d_B{ }^j}(l)\right)}{N(A) \times N(B)}
\end{aligned}
$}
\]
This kernel assigns a constant score $\alpha$ to aligned child nodes, while $\nu$ serves as a decay factor to prevent excessive growth in the final score. Here, $c_{n_i}$ denotes the set of child nodes of $n_i$, and $a_i$ refers to an element within this set.
\[
\scalebox{0.85}{$
\begin{aligned}
 K_c\left(n_i, n_j\right)=
 \begin{cases}
     \alpha s\left(n_i, n_j\right)+\nu K_c\left(a_i, b_j\right)  &  \forall a_i \in c_{n_i} \\
      \text { and } \forall b_j \in c_{n_j} \text { if }  d_i=d_j \\
     \qquad \text { and } t_i=t_j\\
     0 & otherwise
 \end{cases}
 \end{aligned}
$}
\]

Finally, the TABK and MSK can be combined to form a Composite Kernel (CK), where the parameters $\beta$ and $\delta$ determine the respective contributions of each kernel.
\[
\scalebox{0.85}{$
\begin{aligned}
     C K(A, B)=\beta. T A B K(A, B) + \delta.M S K(A, B)
\end{aligned}
$}
\]

To the best of our knowledge, these carefully constructed, step-by-step kernels represent the first comprehensive effort to intricately model the similarities within dependency structures. A particularly notable feature is its consideration given to the weights assigned to each node. Moreover, these kernels not only capture the structural nuances of dependencies but also take into account the comparison of labels.

However, there are still certain limitations associated with these kernels. One major concern is the use of tf-idf values to update the weights of dependencies. This is due to the inherent nature of tf-idf, which measures the significance of a token in distinguishing or classifying a document within a collection. As a result, the effectiveness of this weighting mechanism in accurately emphasizing the importance of keywords and their related dependencies remains uncertain. Additionally, while the Matching Subtrees Kernel (MSK) effectively aligns subgraphs, it falls short in fully capturing the influence of type matches during the comparison of two substructures.

Our approach builds on this work. Addressing both their strengths and limitations, this paper presents a novel framework for quantifying cross-lingual syntactic similarities and injecting them to pretrained LMs.
% \subsection{Injection of UD}
      
\label{related}
\section{Visualizing Experimental Results}
\label{sec:appendix-figure}
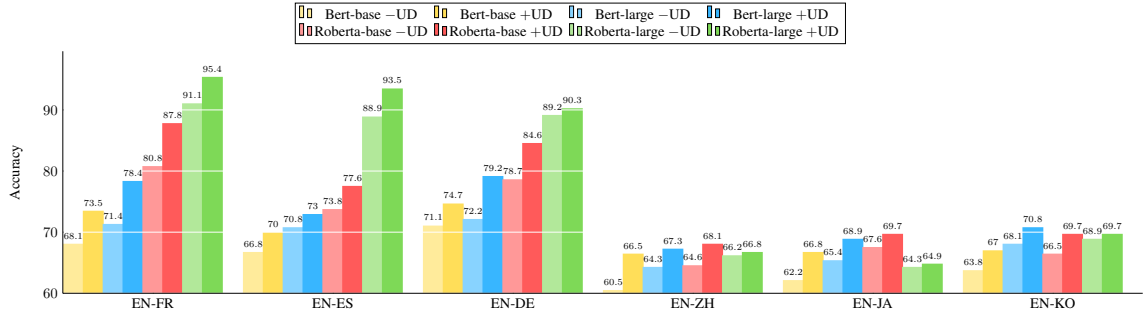
\begin{figure*}[htbp]
% \subfloat[Chinese]{
% \begin{minipage}[t]{0.45\textwidth}
  \centering
  \scalebox{0.5}{
  \begin{tikzpicture}
    \centering
      \pgfplotsset{bar7/.style={mark=no markers,  bar shift=-\pgfplotbarwidth/2*7.6}}
    \pgfplotsset{bar8/.style={mark=no markers, bar shift=-\pgfplotbarwidth/2*5.7}}
    \pgfplotsset{bar1/.style={mark=no markers,  bar shift=-\pgfplotbarwidth/2*3.8}}
    \pgfplotsset{bar2/.style={mark=no markers, bar shift=-\pgfplotbarwidth/2*1.9}}
    \pgfplotsset{bar3/.style={mark=no markers,  bar shift=-\pgfplotbarwidth/2*0.0}}
    \pgfplotsset{bar4/.style={mark=no markers, bar shift=\pgfplotbarwidth/2*1.9}}
    \pgfplotsset{bar5/.style={mark=no markers,  bar shift=\pgfplotbarwidth/2*3.8}}
    \pgfplotsset{bar6/.style={mark=no markers, bar shift=\pgfplotbarwidth/2*5.7}}

    \begin{axis}[
      ybar=2.5cm, axis on top,
      y tick label style={/pgf/number format/.cd},
      title={},
      height=8cm, width=30cm,
      bar width=0.55cm,
      ymajorgrids, tick align=inside,
      major grid style={draw=white},
      enlarge y limits={value=.1,upper},
      ymin=60.00, ymax=96.00,
      bar shift={\pgfplotbarwidth},
      axis x line*=bottom,
      axis y line*=left,
      % y axis line style={opacity=0},
      tickwidth=0pt,
      enlarge x limits=0.1,
      legend style={
        at={(0.47, 1.2)},
        anchor=north,
        legend columns=4,
        /tikz/every even column/.append style={column sep=0.1cm}
      },
      ylabel={Accuracy},
      % ylabel=Left label,
      symbolic x coords={
       EN-FR, EN-ES, EN-DE, EN-ZH, EN-JA, EN-KO 
      },
      xtick=data,
      every node near coord/.append style={font=\scriptsize},
      nodes near coords={
        \pgfmathprintnumber[precision=1]{\pgfplotspointmeta}
      }
      ]
      \addplot [bar7, draw=none, fill=baryellow!60] coordinates {
        (EN-FR, 68.11)
        (EN-ES, 66.76)
        (EN-DE, 71.08)
        (EN-ZH, 60.54 )
        (EN-JA, 62.16 )
        (EN-KO, 63.78 )
      };\label{plot:BiLSTM-tust.zh}
      \addlegendentry{-ud bert-base}
      \addplot [bar8, draw=none, fill=baryellow] coordinates {
        (EN-FR, 73.51)
        (EN-ES, 70.00)
        (EN-DE, 74.69)
        (EN-ZH, 66.49)
        (EN-JA, 66.76)
        (EN-KO, 67.03)
      };\label{plot:BiLSTM-tust.zh}
      \addlegendentry{+ud bert-base}

      \addplot [bar1, draw=none, fill=barblue!60] coordinates {
        (EN-FR, 71.35)
        (EN-ES, 70.81)
        (EN-DE, 72.16)
        (EN-ZH, 64.32)
        (EN-JA, 65.41)
        (EN-KO, 68.11 )
      };\label{plot:BiLSTM-tust.zh}
      \addlegendentry{-ud bert-large}

      \addplot [bar2, draw=none, fill=barblue] coordinates {
        (EN-FR, 78.38 )
        (EN-ES, 72.97 )
        (EN-DE, 79.19 )
         (EN-ZH, 67.30)
          (EN-JA, 68.92)
        (EN-KO, 70.81)
      };\label{plot:+CRF-tust.zh}
      \addlegendentry{+ud bert-large}

      \addplot [bar3, draw=none, fill=barred!50] coordinates {
        (EN-FR, 80.81 )
        (EN-ES, 73.78 )
        (EN-DE, 78.65 )
         (EN-ZH, 64.59 )
          (EN-JA, 67.57 )
        (EN-KO, 66.49 )
      };\label{plot:+ELMo-tust.zh}
      \addlegendentry{-ud roberta}

      \addplot [bar4, draw=none, fill=barred!80] coordinates {
        (EN-FR, 87.84 )
        (EN-ES, 77.57 )
        (EN-DE, 84.59 )
        (EN-ZH, 68.11 )
         (EN-JA, 69.73 )
        (EN-KO, 69.73 )
      };\label{plot:+ELMo+CRF-tust.zh}
      \addlegendentry{+ug f1}

      \addplot [bar5, draw=none, fill=bargreen!60] coordinates {
        (EN-FR, 91.08 )
        (EN-ES, 88.92 )
        (EN-DE, 89.18 )
        (EN-ZH, 66.22 )
         (EN-JA, 64.32 )
        (EN-KO, 68.92 )
      };\label{plot:+BERT-tust.zh}
      \addlegendentry{+BERT}

      \addplot [bar6, draw=none, fill=bargreen] coordinates {
        (EN-FR, 95.41 )
        (EN-ES, 93.51 )
        (EN-DE, 90.27 )
        (EN-ZH, 66.76 )
         (EN-JA, 64.85)
        (EN-KO, 69.73 )
      };
      \label{plot:+BERT+CRF-tust.zh}
      \addlegendentry{+BERT+CRF}
      \legend{Bert-base $-$UD, Bert-base $+$UD, Bert-large $-$UD, Bert-large $+$UD, Roberta-base $-$UD, Roberta-base $+$UD, Roberta-large $-$UD, Roberta-large $+$UD}
    \end{axis}

  \end{tikzpicture}}
\caption{\label{fig:bar-english}Accuracy of different pretrained language models in the cross-lingual adversarial PI task.}
\end{figure*}

\begin{figure*}[htbp]
% \subfloat[Chinese]{
% \begin{minipage}[t]{0.45\textwidth}
  \centering
  \scalebox{0.5}{
  \begin{tikzpicture}
    \centering
     \pgfplotsset{bar7/.style={mark=no markers,  bar shift=-\pgfplotbarwidth/2*7.6}}
    \pgfplotsset{bar8/.style={mark=no markers, bar shift=-\pgfplotbarwidth/2*5.7}}
    \pgfplotsset{bar1/.style={mark=no markers,  bar shift=-\pgfplotbarwidth/2*3.8}}
    \pgfplotsset{bar2/.style={mark=no markers, bar shift=-\pgfplotbarwidth/2*1.9}}
    \pgfplotsset{bar3/.style={mark=no markers,  bar shift=-\pgfplotbarwidth/2*0.0}}
    \pgfplotsset{bar4/.style={mark=no markers, bar shift=\pgfplotbarwidth/2*1.9}}
    \pgfplotsset{bar5/.style={mark=no markers,  bar shift=\pgfplotbarwidth/2*3.8}}
    \pgfplotsset{bar6/.style={mark=no markers, bar shift=\pgfplotbarwidth/2*5.7}}

    \begin{axis}[
      ybar=2.5cm, axis on top,
      y tick label style={/pgf/number format/.cd},
      title={},
      height=8cm, width=30cm,
      bar width=0.55cm,
      ymajorgrids, tick align=inside,
      major grid style={draw=white},
      enlarge y limits={value=.1,upper},
      ymin=45.00, ymax=95.00,
      bar shift={\pgfplotbarwidth},
      axis x line*=bottom,
      axis y line*=left,
      % y axis line style={opacity=0},
      tickwidth=0pt,
      enlarge x limits=0.1,
      legend style={
        at={(0.47, 1.2)},
        anchor=north,
        legend columns=4,
        /tikz/every even column/.append style={column sep=0.1cm}
      },
      ylabel={$F_1$ scores},
      % ylabel=Left label,
      symbolic x coords={
       EN-FR, EN-ES, EN-DE, EN-ZH, EN-JA, EN-KO 
      },
      xtick=data,
      every node near coord/.append style={font=\scriptsize},
      nodes near coords={
        \pgfmathprintnumber[precision=1]{\pgfplotspointmeta}
      }
      ]

    \addplot [bar7, draw=none, fill=baryellow!60] coordinates {
        (EN-FR, 64.02)
        (EN-ES, 64.76)
        (EN-DE, 68.06)
        (EN-ZH, 49.66)
        (EN-JA, 59.77)
        (EN-KO, 55.92)
      };\label{plot:BiLSTM-tust.zh}
      \addlegendentry{-ud bert-base}
      \addplot [bar8, draw=none, fill=baryellow] coordinates {
        (EN-FR, 70.83)
        (EN-ES, 68.56)
        (EN-DE, 72.67)
        (EN-ZH, 58.94 )
        (EN-JA, 60.70 )
        (EN-KO, 64.74 )
      };\label{plot:BiLSTM-tust.zh}
      \addlegendentry{+ud bert-base}
      
      \addplot [bar1, draw=none, fill=barblue!60] coordinates {
        (EN-FR, 61.59)
        (EN-ES, 64.71)
        (EN-DE, 65.32)
        (EN-ZH, 46.34 )
        (EN-JA, 55.56 )
        (EN-KO, 59.86 )
      };\label{plot:BiLSTM-tust.zh}
      \addlegendentry{-ud bert-large}

      \addplot [bar2, draw=none, fill=barblue] coordinates {
        (EN-FR, 75.16 )
        (EN-ES, 66.44 )
        (EN-DE, 76.01 )
         (EN-ZH, 61.09 )
          (EN-JA, 60.48 )
        (EN-KO, 65.16 )
      };\label{plot:+CRF-tust.zh}
      \addlegendentry{+ud bert-large}

      \addplot [bar3, draw=none, fill=barred!50] coordinates {
        (EN-FR, 79.42 )
        (EN-ES, 73.13 )
        (EN-DE, 75.84 )
         (EN-ZH, 52.01 )
          (EN-JA, 59.46 )
        (EN-KO, 59.21 )
      };\label{plot:+ELMo-tust.zh}
      \addlegendentry{-ud roberta-base}

      \addplot [bar4, draw=none, fill=barred!80] coordinates {
        (EN-FR, 86.57 )
        (EN-ES, 77.75 )
        (EN-DE, 82.57 )
        (EN-ZH, 61.69 )
         (EN-JA, 60.56 )
        (EN-KO, 64.33 )
      };\label{plot:+ELMo+CRF-tust.zh}
      \addlegendentry{+ud roberta-base}

      \addplot [bar5, draw=none, fill=bargreen!60] coordinates {
        (EN-FR, 90.21 )
        (EN-ES, 87.91 )
        (EN-DE, 87.65 )
        (EN-ZH, 59.81 )
        (EN-JA, 52.52 )
        (EN-KO, 62.78 )
      };\label{plot:+BERT-tust.zh}
      \addlegendentry{+BERT}

      \addplot [bar6, draw=none, fill=bargreen] coordinates {
       (EN-FR, 94.64 )
       (EN-ES, 92.64 )
        (EN-DE, 89.02 )
         (EN-ZH, 61.20 )
          (EN-JA, 61.64 )
        (EN-KO, 68.18 )
      };
      \label{plot:+BERT+CRF-tust.zh}
      \addlegendentry{+BERT+CRF}
      \legend{Bert-base $-$UD, Bert-base $+$UD, Bert-large $-$UD, Bert-large $+$UD, Roberta-base $-$UD, Roberta-base $+$UD, Roberta-large $-$UD, Roberta-large $+$UD}
    \end{axis}

  \end{tikzpicture}}
\caption{\label{fig:bar-f1}$F_1$ scores of different pretrained language models in the cross-lingual adversarial PI task.}
\end{figure*}
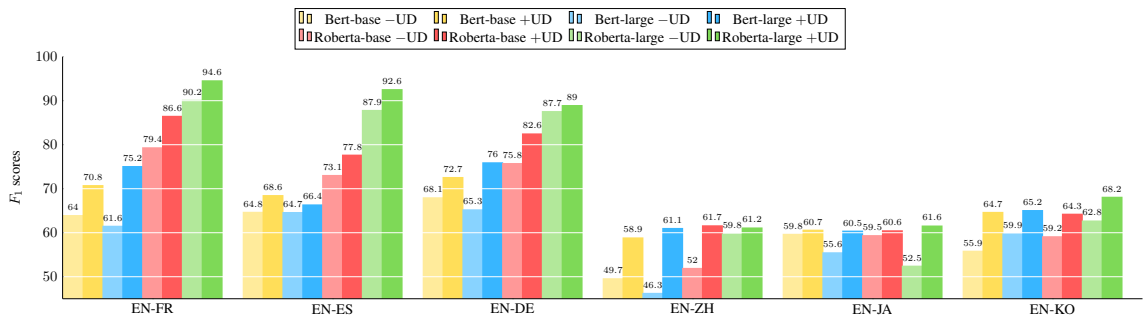

Figure \ref{fig:bar-english} presents accuracies of different pretrained models in identifying the cross-lingual adversarial paraphrases before and after integrating Universal Dependencies, while Figure \ref{fig:bar-f1} shows their $F_1$ scores. 
\section{Implementation Details}
\label{sec:appendix-detail}
For PLMs, the experimental settings are as follows: (i) all pretrained models are trained with a batch size of 16;
(ii) the max length for text encoding is set to 128;
(ii) the dropout rate is set to 0.1; 
(iii) learning rates are selected from 
{1e-5, 2e-5, 8e-6}; 
(iv) the warm-up rate is set to 0.1; 
(v) L2 weight decay is set to 1e-8; 
(vi) the constants $\theta$ and $\beta$ are set to 1.5 and 0.2 respectively.

For the LLM, we fine-tune its linear layers using QLoRA \citep{dettmers2023qloraefficientfinetuningquantized}. We adopt the same hyperparameters for LoRA rank ($r$), LoRA alpha ($\alpha$), and dropout ($d$) as those used in \citet{ruan-etal-2024-large}.
% All pretrained models were trained on V100 GPUs.

\end{document}